# Your Device May Know You Better Than You Know Yourself- Continuous Authentication on novel dataset Using machine learning

Pedro Gomes do Nascimento, Pidge Witiak, Tucker MacCallum, Zachary Winterfeldt, Rushit Dave

Department of Computer Information Science, Minnesota State University, Mankato, Mankato, USA

pedro.gomesdonascimento@mnsu.edu

pidge.witiak@mnsu.edu

tucker.maccallum@mnsu.edu

zachary.winterfeldt@mnsu.edu

rushit.dave@mnsu.edu

## Abstract

*This research aims to further understanding in the field of continuous authentication using behavioural biometrics. We are contributing a novel dataset that encompasses the gesture data of 15 users playing Minecraft with a Samsung Tablet, each for a duration of 15 minutes. Utilizing this dataset, we employed machine learning (ML) binary classifiers, being Random Forest (RF), K-Nearest Neighbors (KNN), and Support Vector Classifier (SVC), to determine the authenticity of specific user actions. Our most robust model was SVC, which achieved an average accuracy of approximately 90%, demonstrating that touch dynamics can effectively distinguish users. However, further studies are needed to make it viable option for authentication systems. You can access our dataset at the following link: https://github.com/AuthenTech2023/authentech-repo*

# 1. INTRODUCTION

The current authentication methods, which are primarily implemented at entry points, can be problematic in numerous scenarios. Once a device is unlocked, it remains unlocked. The device can also be vulnerable to smudge attacks and keyloggers. In a world of ever-advancing technology, security of our devices is vital, and a proposed solution for authentication is the implementation of continuous authentication methods. Continuous authentication relies on the idea that a device could learn the user's behavioural biometrics patterns and identify whether an impostor or the actual user is using the device. Within this paper, we investigate continuous authentication methods rooted in machine learning. Our research is grounded in touchscreen data acquired from 15 volunteers playing the popular game Minecraft, from which we extracted essential features. These features served as the foundation for training and assessing our machine learning models. The three types of machine learning models we used were: RF, KNN and SVC. In this paper, we make the following contributions:

1. We created a public touch dynamics dataset that records the actions of 15 users who played Minecraft on an Android device. The dataset is available at: https://github.com/AuthenTech2023/authentech-repo.
2. We trained and tested three models: KNN, SVC and RF. We then compared our outcome to results from previous works.

This paper is organized as follows: Section 2 reviews the related literature on this field and highlights the research gap. Section 3 describes the methodology, including the data collection and the models employed. Section 4 presents the results. Section 5 discusses the findings and compares them with existing studies. Section 6 acknowledges the limitations of the research and suggests directions for future work. Section 7 concludes the paper and summarizes the main contributions.

# 2. LITERATURE REVIEW

User authentication through touchscreen gestures has gained significant attention in recent years, fueled by the proliferation of smartphones and the need for robust security measures. The existing body of literature offers valuable insights into various models and methodologies employed to authenticate users based on their touch behavior. However, a notable gap in the literature lies in the limited exploration of individual user-centric authentication as opposed to the prevalent focus on user-versus-imposter scenarios.

Several studies have addressed the broader context of touchscreen authentication, providing foundational knowledge of sensors commonly found in smartphones [1]. Achieving an 87% accuracy using a Siamese Neural Network architecture, this work sets the stage for more nuanced investigations [1]. Others have leveraged Random Forest to extract features and MultiLayer Perceptron for user authentication, achieving an impressive accuracy of 95.96% [2]. The exploration of different machine learning models has been a recurrent theme, with varied approaches such as K-means clusters and SVM [3], one-class SVM [4], and fusion models like SVM and GMM adapted from UBM [5].

However, the literature primarily revolves around generic touch tasks or predefined gestures without considering individual user nuances during free tasks. In contrast, this paper chooses an approach focusing on authenticating individual users within a constrained yet task-free setting. The choice of Minecraft as the application for data collection proves strategic, as it demands users to perform a diverse range of touchscreen gestures, which can garner better results [6][7][8]. The popularity and intuitiveness of Minecraft further enhance the applicability of the proposed solution.

Our selection of Support Vector Classification (SVC), k-Nearest Neighbors (KNN), and Random Forest as the machine learning models is informed by the success observed in previous studies [9][10][11]. Notably, KNN and Random Forest have demonstrated promising results, with average AUC values of 0.9091 and 0.9698, respectively [12]. The rationale behind this choice lies in the pursuit of models that showcase efficacy in touchscreen-based user authentication.

To contribute to the existing knowledge, we introduce a novel testing methodology by individually evaluating each user's touchscreen data rather than adopting the conventional user-versus-imposter paradigm. The 30-70 test split allows for comprehensive evaluation, ensuring a robust understanding of each model's performance for each user. Moreover, our decision to make the touchscreen data publicly available aims to foster collaboration and further advancements in the field.

Drawing inspiration from prior research [10][11], we carefully extracted features for model training, considering their efficacy in producing accurate and reliable results. Emphasizing adaptability to real-world applications, our work strives to bridge the gap between task-specific accuracy, which can lead to better results [13], and practical usability.

[14] presents a practical scheme for continuous user authentication on smartphones, incorporating a risk score mechanism. [15] proposed a global model eliminating the need for retraining on new users, addressing the challenge of computing resource constraints on mobile devices. These paper provide insights into the scalability and applicability of touchscreen authentication solutions in practical settings.

In summary, this literature review positions our study within the broader landscape of touchscreen-based user authentication, highlighting the unique contributions of our approach. By focusing on individual user-centric authentication within the context of a popular application like Minecraft, and employing machine learning models with proven success, we aim to advance the understanding of touchscreen authentication and pave the way for practical, user-centric solutions.

## 3. METHODS

### 3.1. Data Collection

This research involved human subjects and required ethical training and approval. We completed the Collaborative Institutional Training Initiative (CITI) program, which covers various topics such as history and ethical principles, informed consent, privacy and confidentiality, students in research, and defining research with human subjects. We also obtained approval from the Institutional Review Board (IRB) to conduct the data collection on the Mankato State University campus with student participants.

### 3.2. Dataset

| Metrics | Description |
| --- | --- |
| Timestamp | Time of recorded touch event |
| X | X coordinate |
| Y | Y coordinate |
| BTN_TOUCH | Down - screen was just touched. Held - finger is on screen. Up - touch was just released. |

| TOUCH_MAJOR | Length of the major axis of the contact. |
|---|---|
| TOUCH_MINOR | Length of the minor axis of the contact. |
| TRACKING_ID | Identifies an initiated contact throughout its life cycle. |
| PRESSURE | The pressure on the contact area. |
| FINGER | Whenever there is a second finger on the screen it will have the value of "1". Default value is "0". |

Table 1 – Metrics collected and their description.

Data collection was performed using a python script that utilized the Android Debug Bridge (ADB) tool to access the touch screen metrics of the device. Our dataset consists of raw touch dynamics data from fifteen volunteers that played Minecraft for fifteen minutes on a Samsung A8 tablet. The metrics extracted from the device, presented in Table 1, include x and y coordinates, length and width of fingertip, touch pressure, when the screen is touched and released, and finger ids. The python script used for the data collection can be found at: https://github.com/AuthenTech2023/authentech-repo.

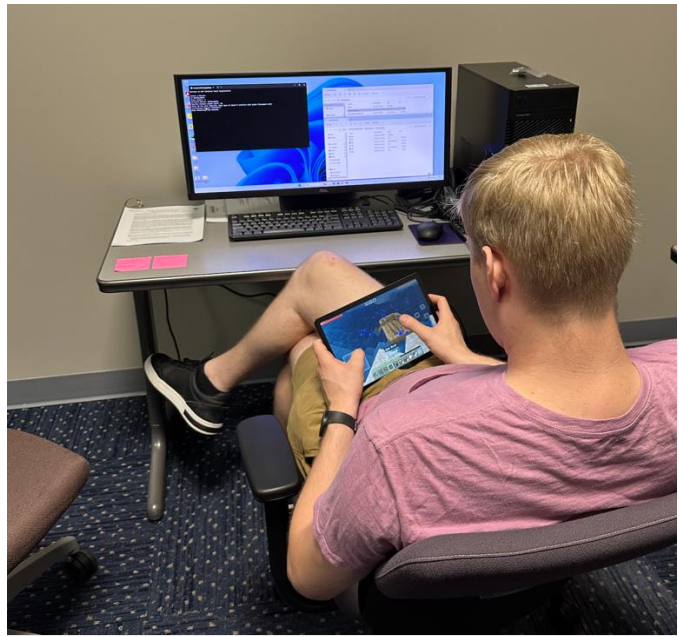

Figure 1-Volunteer in data collection session.

### 3.3. Data Cleaning and Processing

Data filtering techniques were applied to improve the quality and reliability of the dataset. Specifically, rows with default values of "-420" in either the X or Y coordinates were excluded to enhance the integrity of the data. The dataset was sorted by finger ID and timestamp to facilitate meaningful analysis. We converted all string values into numerical representations to ensure we could perform proper calculations with the data. Any rows containing missing (NaN) values were removed to ensure data completeness.

In addition to these filtering procedures, the numerical columns underwent a standardization process. This involved making the numbers more balanced and consistent. We used a tool from Sklearn called StandardScaler() that changes the numbers so that their average is 0 and their variation is 1. This makes it easier to compare them and see patterns. This standardization not only aids in achieving uniformity across features but also contributes to the interpretability of subsequent analyses.

These sequential steps, including data exclusion, sorting, conversion, and standardization, collectively contribute to the cleansing and enhancement of the dataset.

### 3.4. Feature Extraction

In the process of feature extraction from our cleaned and pre-processed dataset, we specifically selected the following key features: X_Speed and Y_Speed, representing the touch's instantaneous velocities along the X and Y axes; Speed, an aggregate speed measure derived from the X and Y individual components; X_Acceleration and Y_Acceleration, denoting the instantaneous accelerations along the horizontal and vertical directions; Acceleration, a comprehensive measure derived from the X and Y individual accelerations; Jerk, representing the rate of change of acceleration; Path_Tangent, capturing the touch's path angle; and Ang_V (Angular Velocity), providing insights into the touch's rotational dynamics.

Figure 2 - User 9's touch features

## 4. METHODOLOGIES

In implementing our chosen ML models, we elected to train each model on all 15 users, and then tested the models on each user's data individually. This served as a challenge with accuracy, because it increased variability between users' results when the same model was deployed. 70% of each user's data was used to train the model, then the remaining 30% of each user's data was used for testing. The end-to-end process is shown below in Figure 3. To sum it up: We started by collecting raw data from 15 subjects, then process the data into the features we extracted. Then, we performed our train-test split, trained a model, tested it, and obtained our results. This process was followed for all three models.

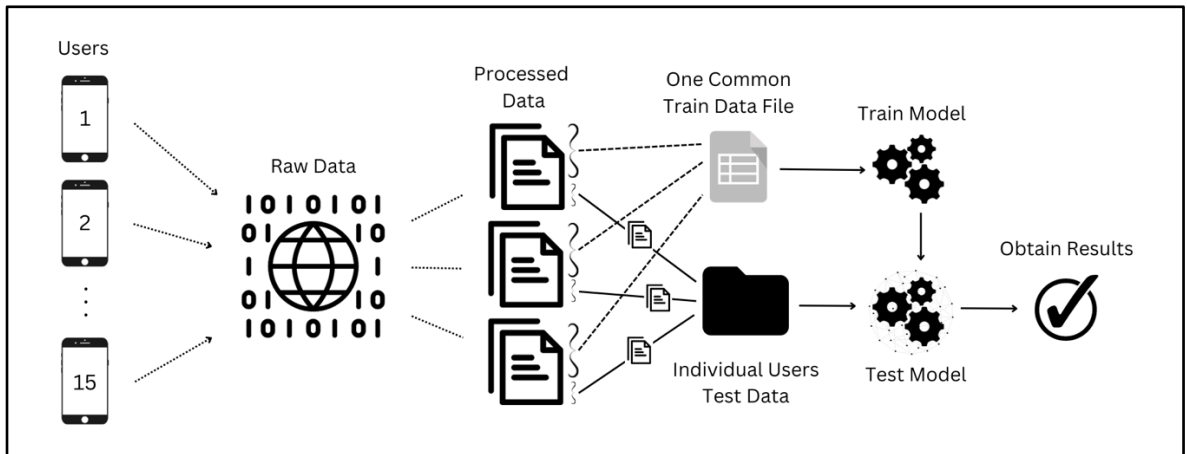

Figure 3 – End-to-end process of the experiment.

## 4.5. The Models

### 4.5.1. K-Nearest-Neighbors

The KNN model classifies users based on the proximity of a data point to other classified data points. When training the model, labelled data points are plotted on a n-dimensional graph where n is the number of features in the data (one axis for each feature). Users are then classified according to the k closest data points, where k is an integer given to the model along with the testing data. If the k value is 3 and the 3 closest data points are of class 1, 1, and 2, the data point will be classified as class 1. When determining the best k value for our data set, we employed the use of the "Elbow Method". The elbow method is a means of determining appropriate k values by running the model with a range of k values and plotting the error rate against the k values. From this graph the k value that looks to be at the moment of flattening error rates is deemed to be the best fit. To determine our k value, we used this method for each user's dataset and chose the k value that was best fit for the most users (k=2).

```python
def knn_model(auth_user, k=4):

    # IMPORT TRAIN AND TEST DATA
    X_train = pd.read_csv('processed-feature-data/training-data/X_training_data.csv')
    X_test = pd.read_csv('processed-feature-data/testing-data/X_testing_data_user' + str(auth_user) + '.csv')
    y_train = pd.read_csv('processed-feature-data/training-data/y_training_data.csv')
    y_test = pd.read_csv('processed-feature-data/testing-data/y_testing_data_user' + str(auth_user) + '.csv')

    # FIT THE MODEL
    knn = KNeighborsClassifier(n_neighbors=k)  # n_neighbors should be adjusted to best value
    knn.fit(X_train.values, y_train.values.ravel())

    # TEST MODEL
    pred = knn.predict(X_test.values)

    # EVALUATE METRICS
    conf_matrix = confusion_matrix(y_test, pred, labels=[1,2,3,4,5,6,7,8,9,10,11,12,13,14,15])
    class_report = classification_report(y_test, pred)
    pred = list(pred)

    # print to file (Necessary for confusion-matrix-display.py)
    output_directory = os.path.join("model-outputs", "knn", 'kvalue' + str(k))
    os.makedirs(output_directory, exist_ok=True)
    with open('model-outputs/knn/kvalue' + str(k) + '/user' + str(auth_user) + '_confusion_matrix.txt', mode="w") as f:
        f.write(str(conf_matrix))
    with open('model-outputs/knn/kvalue' + str(k) + '/user' + str(auth_user) + '_classification_report.txt', mode="w") as f:
        f.write(str(class_report))
    with open('model-outputs/knn/kvalue' + str(k) + '/user' + str(auth_user) + '_pred.txt', mode="w") as f:
        f.write(str(pred))

    # print to std
    print(conf_matrix)
    print(class_report)

    # RETURNS FOR ELBOW METHOD
    return pred, y_test.values.ravel()

if __name__ == '__main__':
    # authentic_user = 1
    k = 2
    # knn_model(authentic_user, k)

    # multiple users
    for authentic_user in range(1,16):
        print(f'user {authentic_user} start')
        knn_model(authentic_user, k)
```

Figure 4 – kNN Python code

### 4.5.2. Random Forest

The RF algorithm is based on the decision trees algorithm, which looks at all of the features, and sees when certain classifications are made based off of certain features paired together. In respect to this, a random forest uses multiple decision trees with random feature selection in attempts to reduce overfitting. We implemented the random forest algorithm using scikit-learn's RandomForestClassifier class in Python. Normalization, scaling, testing different features, and running a grid search were all undergone in efforts to attain the best results. The grid search revealed to us the optimal hyperparameters to use for each user. With this information, we recorded the mean and mode of these hyperparameters for all users. We even experimented with choosing different hyperparameters including max_depth, max_features, min_samples_leaf, and min_samples_split. The final hyperparameters we ended up utilizing were min_samples_leaf=3 and max_depth=6.

```python
X_train = pd.read_csv('processed-feature-data/training-data/X_training_data.csv')
y_train = pd.read_csv('processed-feature-data/training-data/y_training_data.csv')

# param_grid = {'max_depth':[7, 9, 11, 13], 'max_features':['sqrt','log2'],
#              'min_samples_leaf':[2,3,4,5], 'min_samples_split':[7,8,9,10]}
# print('param grid set.')
#
# grid = GridSearchCV(RandomForestClassifier(), param_grid, verbose=2)
# print('Fitting...')
# grid.fit(X_train,y_train.values.ravel())
# print("Done.")
# print('Grid search complete')

# Found dropping these metrics to be beneficial
X_train = X_train.drop(['Timestamp'], axis=1)

rfc = RandomForestClassifier(n_estimators=300, max_depth=7, min_samples_leaf=3)
print('fitting model')
rfc.fit(X_train, y_train.values.ravel())
print('\nModel fitted.\n')

# todo: Add max_features parameter once you figure out how it works

# todo: Also add max_features implementation

def random_forest_model(auth_user, max_depth=7, min_samples_leaf=3):
    X_test = pd.read_csv('processed-feature-data/testing-data/X_testing_data_user' + str(auth_user) + '.csv')
    y_test = pd.read_csv('processed-feature-data/testing-data/y_testing_data_user' + str(auth_user) + '.csv')

    X_test = X_test.drop(['Timestamp'], axis=1)

    pred = rfc.predict(X_test)

    # EVALUATE METRICS
    conf_matrix = confusion_matrix(y_test, pred, labels=[1, 2, 3, 4, 5, 6, 7, 8, 9, 10, 11, 12, 13, 14, 15])
    class_report = classification_report(y_test, pred)
    pred = list(pred)

    # print to file (Necessary for confusion-matrix-display.py)
    output_directory = os.path.join("model-outputs", "random-forest", 'max-depth-' + str(max_depth))
    os.makedirs(output_directory, exist_ok=True)
    with open('model-outputs/random-forest/max-depth-' + str(max_depth) + '/user' + str(auth_user) + '_confusion_matrix.txt', mode="w") as f:
        f.write(str(conf_matrix))
    with open('model-outputs/random-forest/max-depth-' + str(max_depth) + '/user' + str(auth_user) + '_classification_report.txt',mode="w") as f:
        f.write(str(class_report))
    with open('model-outputs/random-forest/max-depth-' + str(max_depth) + '/user' + str(auth_user) + '_pred.txt', mode="w") as f:
        f.write(str(pred))

    return conf_matrix
```

Figure 5 – Random Forest Python code

### 4.5.3. Support Vector Classifier

The SVC works by using the training data to find the most optimal "support vector" to split the data to categorize one class versus another. It then uses the same vector to make classifications with the test data. We used scikit-learn's SVC library and its methods to deploy our own model in Python. It's worth noting that this algorithm was extremely time consuming to execute, especially when initially fitting the model. Some possible hyperparameters that could've been changed were the C and gamma values, but no hyperparameter tuning was necessary with our data and model.

```python
27  X_train = pd.read_csv('processed-feature-data/training-data/X_training_data.csv')
28  y_train = pd.read_csv('processed-feature-data/training-data/y_training_data.csv')
29  X_train = X_train.drop(['Timestamp'], axis=1)
30  svc = SVC()
31  print('Model fitting')
32  svc.fit(X_train, y_train.values.ravel())
33  print('Model fit')
34
35  def SVC_model(auth_user):
36
37      # todo: Make the deep copy, make sure data is in RAM
38
39      X_test = pd.read_csv('processed-feature-data/testing-data/X_testing_data_user' + str(auth_user) + '.csv')
40      y_test = pd.read_csv('processed-feature-data/testing-data/y_testing_data_user' + str(auth_user) + '.csv')
41
42      X_test = X_test.drop(['Timestamp'], axis=1)
43
44      pred = svc.predict(X_test)
45
46      # EVALUATE METRICS
47      conf_matrix = confusion_matrix(y_test, pred, labels=[1, 2, 3, 4, 5, 6, 7, 8, 9, 10, 11, 12, 13, 14, 15])
48      class_report = classification_report(y_test, pred)
49      pred = list(pred)
50
51      # print to file (Necessary for confusion-matrix-display.py)
52      output_directory = os.path.join("model-outputs", "svc")
53      os.makedirs(output_directory, exist_ok=True)
54      with open('model-outputs/svc/' + '/user' + str(auth_user) + '_confusion_matrix.txt', mode="w") as f:
55          f.write(str(conf_matrix))
56      with open('model-outputs/svc/' + '/user' + str(auth_user) + '_classification_report.txt', mode="w") as f:
57          f.write(str(class_report))
58      with open('model-outputs/svc/' + '/user' + str(auth_user) + '_pred.txt', mode="w") as f:
59          f.write(str(pred))
60
61      return conf_matrix
62
63      # Don't need
64      # print(classification_report(y_test, predictions))
65
66
67  if __name__ == "__main__":
68      for i in range(1, 16):
69          print("User " + str(i) + " results:")
70          print(SVC_model(i))
71
72          cm = SVC_model(i)
73          tp = cm[i - 1][i - 1]
74          # print(tp)
75          fn = sum(cm[i - 1]) - tp
76          # print(fn)
```

Figure 6 – SVC Python code

## 5. RESULTS

Most of the evaluations that we made came from the results of true positives, false positives, true negatives, and false negatives. These can be seen below in Figure 4. In the context of this study, a true positive (TP) is defined as when an authentic user is classified as themself, a true negative (TN) is when an impostor is correctly classified, a false positive (FP) is when an impostor is classified as the authentic user, and a false negative (FN) is when an authentic user is classified as an impostor.

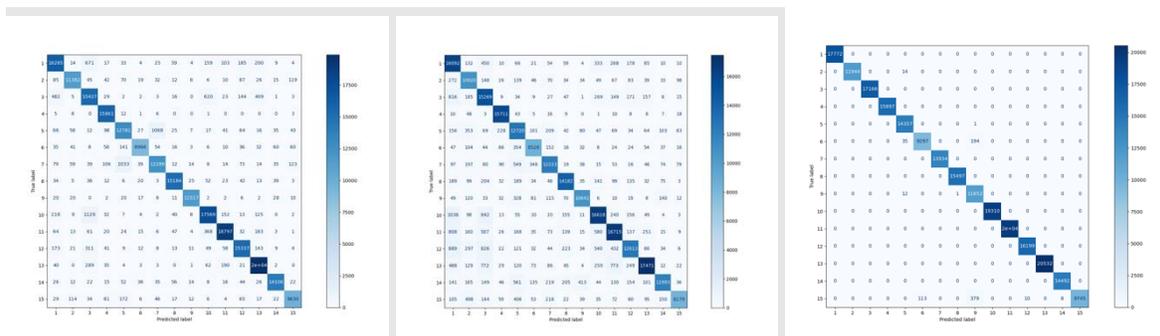

Figure 3 – 2x2 matrix of how TP, FP, FN, and TN are calculated. Image provided by https://medium.com/@awabmohammedomer/confusion-matrix-b504b8f8e1d1

The evaluating our models, we calculated accuracy, precision, recall, f1-score, and area under the curve (AUC). These equations can be seen in Table 2.

| Evaluation Metric | Formula |
|---|---|
| Precision | $\dfrac{TP}{TP + FP}$ |
| Recall | $\dfrac{TP}{TP + FN}$ |
| F1 | $\dfrac{2 * precision * recall}{precision + recall}$ |
| Accuracy | $\dfrac{TP + TN}{TP + FN + TN + FP}$ |

Table 2 - Formulas for evaluation metrics

Figure 8 – Confusion matrices for k-NN, SVC, and RF models respectively

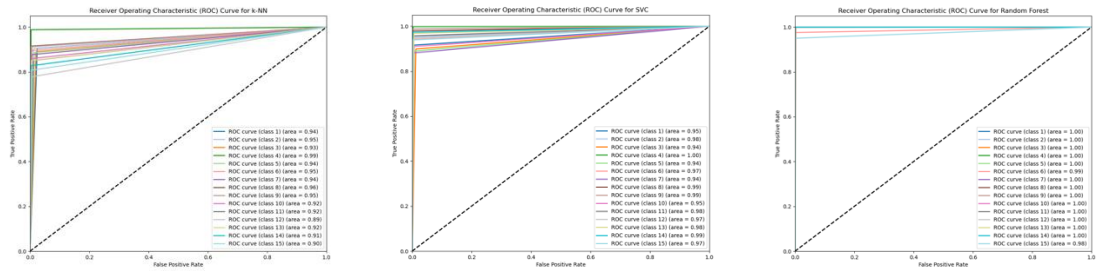

Figure 4 – ROC curves for k-NN, SVC, and RF models respectively

| User | k-NN | SVC | RF |
|---|---|---|---|
| 1 | 70% | 85% | 100% |
| 5 | 73% | 80% | 100% |
| 7 | 80% | 81% | 100% |
| 8 | 86% | 96% | 100% |
| 15 | 78% | 90% | 95% |

Table 3 – Accuracy test for KNN, SVC, and RF respectively.

We can see from the results for all the models in Figures 5,6 and Table 3 that KNN produced above average results, SVC produced exceptional results, and RF produced results that were too good, and likely a product of overfitting: the model learning from the noise of the training data very well but doing poorly with unseen data. This is an assumption we have made based off comparing the near perfect results we obtained to the results of previous similar research. This assumption could be evaluated through plotting loss and accuracy against various tunings of the model and identifying where our model's test results stray from validating results. Many avenues were taken to combat this overfitting. After pairing multiple sets of hyperparameters and manipulating the data even further with little to no improvement in results, we concluded that the random forest model was not the right fit for our data. Due to this, we will be disregarding the results from the random forest model.

| Model | Accuracy | Precision | Recall | F1-score | AUC |
|---|---|---|---|---|---|
| k-NN | 0.78 | 0.88 | 0.88 | 0.88 | 0.934 |
| SVC | 0.9 | 0.94 | 0.94 | 0.95 | ~0.9693 |
| ~~RF~~ | ~~0.99~~ | ~~1.00~~ | ~~0.99~~ | ~~1.00~~ | ~~0.998~~ |

Table 4 - Model evaluation results

The scoring criteria seen in Table 4 are defined as the following:

- Accuracy: The rate of correct classifications out of all classifications, either authentic or impostor.
- Precision: The rate at which the authentic user defined as themselves.

- Recall: Out of all instances of the correct user, the rate that the model was right in classifications.
- F1-score: The harmonic mean of precision and recall.
- AUC: Generated by the area under the receiver operating characteristic curve. Tests results of the model at different thresholds of sensitivity for true positives and false positives.

## 5. DISCUSSION AND ANALYSIS

| Paper | Method | Results |
| --- | --- | --- |
| [1] | Siamese RNN | EER = 13%, Accuracy = 87% |
| [2] | Multilayer Perceptron (MLP) | FRR = 2.55%, FAR = 6.94%, Accuracy = 95.96% |
| [3] | SVC and K-Means | Accuracy = 93.7%, Recall = 86% |
| [16] | SVC | Accuracy = 97.7% |
| [8] | Random Forest | Accuracy = 97%, EER = 3.49% |
| Our Research | KNN and SVC | AUC = 93.4%-97%, Precision = 88%-95%, F1-Score = 88%-94%, Accuracy = 78%-90%, FRR = 6% (SVC) |

Table 5 - A contrastive evaluation of the models we used and other models from previous research papers.

Table 5 provides a comparative analysis of the performance of various machine learning methods used in different research papers and our current study. The methods include Siamese Recurrent Neural Networks (RNN), Multilayer Perceptron (MLP), Support Vector Machines (SVC) and K-Means, Random Forest, and K-Nearest Neighbors (KNN) and Support Vector Classifier (SVC). The performance metrics are accuracy, error rate, and false acceptance rate. The table shows that our most robust model was SVC, which achieved an average accuracy of approximately 90%. This indicates that SVC can effectively distinguish users based on their touch dynamics while playing Minecraft. The table also shows that the other methods, such as RNN, MLP, SVC and K-Means, and Random Forest, also achieved high accuracy rates, ranging from 86% to 97.7%. These results suggest that touch dynamics are a reliable source of behavioural biometrics for continuous authentication. However, the table also reveals some limitations of the methods, such as the high error rate of Siamese RNN (13%) and the high false acceptance rate of Multilayer Perceptron (6.94%). These limitations imply that some methods may be more prone to misclassify users or accept impostors, which can compromise the security of authentication systems. Therefore, further studies are needed to improve the performance and robustness of the methods.

## 6. LIMITATIONS

The main limitation of this study is the small size and diversity of the dataset. We only collected data from 15 users, who were all students from the same university. This may limit the generalizability of our results to other populations and contexts. Moreover, we only used one type of device (Samsung Tablet) and one type of application (Minecraft) to capture the touch dynamics of the users. Different devices and applications may have different effects on the user's behaviour and performance. Therefore, future work should aim to collect more data from a larger and more diverse sample of users, using different devices and applications, to validate and improve our approach.

## 7. CONCLUSION

Our research aimed to further understanding in the field of continuous authentication using behavioural biometrics with the goal of expanding authentication systems past the entry point. Understanding and development of increasingly secure authentication systems is crucial as the quantity of personal data being stored on personal devices continues to increase. The results of our research, along with the contribution of a novel dataset add to the foundation of research in this field by demonstrating the possibility of using touch-screen behavioural biometrics for continuous authentication without clustering data points into segments. The results of our research, along with the research done by other teams suggest that some machine learning models are more fit for this use case than others. As the quantity of personal data stored on touch screen devices continues to increase, the implications of our research and a touchscreen based continuous authentication system become more evident. A system reliable enough to accurately determine the identity of users while they interact with a device would have significant contributions to the field of cyber security and personal data protection.

## REFERENCES

[1]     Acien, A.; Morales, A.; Vera-Rodriguez, R.; Fierrez, J. Smartphone sensors for modeling human-computer interaction: General outlook and research datasets for user authentication. In Proceedings of the 2020 IEEE 44th Annual Computers, Software, and Applications Conference, Madrid, Spain, 13 July 2020. [CrossRef]

[2]     Liang, X.; Zou, F.; Li, L.; Yi, P. Mobile terminal identity authentication system based on behavioral characteristics. Int. J. Distrib. Sens. Netw. 2020, 16, 1550147719899371. [CrossRef]

[3]     Torres, J.; Santos, S.; Alepis, E.; Patsakis, C. Behavioral Biometric Authentication in Android Unlock Patterns through Machine Learning. In Proceedings of the 5th International Conference on Information Systems Security and Privacy (ICISSP), Prague, Czech Republic, 23–25 February 2019. [CrossRef]

[4]     Shi, D.; Tao, D.; Wang, J.; Yao, M.; Wang, Z.; Chen, H.; Helal, S. Fine-grained and context-aware behavioral biometrics for pattern lock on smartphones. Proc. ACM Interact. Mob. Wearable Ubiquitous Technol. 2021, 5, 1–30. [CrossRef]

[5]     J. Cybersecur. Priv. 2023, 3 256 22. Fierrez, J.; Pozo, A.; Martinez-Diaz, M.; Galbally, J.; Morales, A. Benchmarking touchscreen biometrics for mobile authentication. IEEE Trans. Inf. Forensics Secur. 2018, 13, 2720–2733. [CrossRef]

[6]     Lee, J.; Park, S.; Kim, Y.G.; Lee, E.K.; Jo, J. Advanced Authentication Method by Geometric Data Analysis Based on User Behavior and Biometrics for IoT Device with Touchscreen. Electronics 2021, 10, 2583. [CrossRef]

[7]     Antal, M.; Szabó, L.Z. Biometric authentication based on touchscreen swipe patterns. Procedia Technol. 2016, 22, 862–869. [CrossRef]

[8]     Zhang, X.; Zhang, P.; Hu, H. Multimodal continuous user authentication on mobile devices via interaction patterns. Wirel. Commun. Mob. Comput. 2021, 2021, 5677978. [CrossRef]

[9] Samet, S.; Ishraque, M.T.; Ghadamyari, M.; Kakadiya, K.; Mistry, Y.; Nakkabi, Y. TouchMetric: A machine learning based continuous authentication feature testing mobile application. Int. J. Inf. Technol. 2019, 11, 625–631. [CrossRef]

[10] Pelto, Brendan, Mounika Vanamala, and Rushit Dave. "Your Identity is Your Behavior--Continuous User Authentication based on Machine Learning and Touch Dynamics." *arXiv preprint arXiv:2305.09482* (2023).

[11] Z. DeRidder *et al.*, "Continuous User Authentication Using Machine Learning and Multi-finger Mobile Touch Dynamics with a Novel Dataset," *2022 9th International Conference on Soft Computing & Machine Intelligence (ISCMI)*, Toronto, ON, Canada, 2022, pp. 42-46, doi: 10.1109/ISCMI56532.2022.10068450.

[12] Leyfer, K.; Spivak, A. Continuous user authentication by the classification method based on the dynamic touchscreen biometrics. In Proceedings of the 2019 24th Conference of Open Innovations Association (FRUCT), Moscow, Russia, 8 April 2019. [CrossRef]

[13] Li, W.; Meng, W.; Furnell, S. Exploring touch-based behavioral authentication on smartphone email applications in IoT-enabled smart cities. Pattern Recognit. Lett. 2021, 144, 35–41. [CrossRef]

[14] Buriro, A.; Gupta, S.; Yautsiukhin, A.; Crispo, B. Risk-driven behavioral biometric-based one-shot-cum-continuous user authentication scheme. J. Signal Process. Syst. 2021, 93, 989–1006. [CrossRef]

[15] Levi, M.; Hazan, I.; Agmon, N.; Eden, S. Behavioral embedding for continuous user verification in global settings. Comput. Secur. 2022, 119, 102716. [CrossRef]

[16] Miyamoto, N.; Shibata, C.; Kinoshita, T. Authentication by Touch Operation on Smartphone with Support Vector Machine. Int. J. Inf. Secur. Res 2017, 7, 725–733. [CrossRef]

Mallet, J., Pryor, L., Dave, R., Seliya, N., Vanamala, M., & Sowells-Boone, E. (2022, March). Hold On and Swipe: A Touch-Movement Based Continuous Authentication Schema based on Machine Learning. In 2022 Asia Conference on Algorithms, Computing and Machine Learning (CACML) (pp. 442-447). IEEE.

Tee, W. Z., Dave, R., Seliya, N., & Vanamala, M. (2022, March). Human Activity Recognition models using Limited Consumer Device Sensors and Machine Learning. In 2022 Asia Conference on Algorithms, Computing and Machine Learning (CACML) (pp. 456-461). IEEE.

Tee, W. Z., Dave, R., Seliya, J., & Vanamala, M. (2022, May). A Close Look into Human Activity Recognition Models using Deep Learning. In 2022 3rd International Conference on Computing, Networks and Internet of Things (CNIOT) (pp. 201-206). IEEE.

Siddiqui, N., Dave, R., Vanamala, M., & Seliya, N. (2022). Machine and Deep Learning Applications to Mouse Dynamics for Continuous User Authentication. Machine Learning and Knowledge Extraction, 4(2), 502-518.

Mallet, J., Dave, R., Seliya, N., & Vanamala, M. (2022). Using Deep Learning to Detecting Deepfakes. arXiv preprint arXiv:2207.13644.

Pryor, L., Mallet, J., Dave, R., Seliya, N., Vanamala, M., & Boone, E. S. (2022). Evaluation of a User Authentication Schema Using Behavioral Biometrics and Machine Learning. arXiv preprint arXiv:2205.08371.

Deridder, Z., Siddiqui, N., Reither, T., Dave, R., Pelto, B., Seliya, N., & Vanamala, M. (2022). Continuous User Authentication Using Machine Learning and Multi-Finger Mobile Touch Dynamics with a Novel Dataset. arXiv preprint arXiv:2207.13648.

Mason, J., Dave, R., Chatterjee, P., Graham-Allen, I., Esterline, A., & Roy, K. (2020). An investigation of biometric authentication in the healthcare environment. Array, 8, 100042.

Gunn, Dylan J., et al. "Touch-based active cloud authentication using traditional machine learning and LSTM on a distributed tensorflow framework." International Journal of Computational Intelligence and Applications 18.04 (2019): 1950022.

J. Shelton *et al.*, "Palm Print Authentication on a Cloud Platform," *2018 International Conference on Advances in Big Data, Computing and Data Communication Systems (icABCD)*, Durban, South Africa, 2018, pp. 1-6, doi: 10.1109/ICABCD.2018.8465479.

Siddiqui, Nyle, Laura Pryor, and Rushit Dave. "User authentication schemes using machine learning methods—a review." Proceedings of International Conference on Communication and Computational Technologies: ICCCT 2021. Springer Singapore, 2021.

Mallet, Jacob, et al. "Hold on and swipe: a touch-movement based continuous authentication schema based on machine learning." 2022 Asia Conference on Algorithms, Computing and Machine Learning (CACML). IEEE, 2022.

Pryor, Laura, et al. "Machine learning algorithms in user authentication schemes." 2021 International Conference on Electrical, Computer and Energy Technologies (ICECET). IEEE, 2021.